\documentclass[10pt, conference, compsocconf]{IEEEtran}
\usepackage{xspace}
\usepackage{subcaption}
\usepackage{times}
\usepackage{epsfig}
\usepackage{graphicx}
\usepackage{amsmath}
\usepackage{amssymb}
\usepackage{color}
\usepackage{algorithm}
\usepackage{dsfont}
\usepackage{capt-of}
\usepackage{booktabs}
\usepackage[noend]{algpseudocode}
\usepackage[group-separator={,}]{siunitx}

\graphicspath{{Figures/}}
\makeatletter
\DeclareRobustCommand\onedot{\futurelet\@let@token\@onedot}
\def\@onedot{\ifx\@let@token.\else.\null\fi\xspace}
\usepackage{ textcomp }
 
\def\ie{\emph{i.e}\onedot}

\def\etal{\emph{et al}\onedot}

\usepackage{cite}

\ifCLASSINFOpdf
\else
\fi
\usepackage{url}

\let\OLDthebibliography\thebibliography
\renewcommand\thebibliography[1]{
  \OLDthebibliography{#1}
  \setlength{\parskip}{0pt}
  \setlength{\itemsep}{0pt plus 0.3ex}
}

\begin{document}
%
\title{Real-Time Human Motion Capture with Multiple Depth Cameras}


\author{\IEEEauthorblockN{Alireza Shafaei, James J. Little}
\IEEEauthorblockA{Computer Science Department\\
The University of British Columbia\\
Vancouver, Canada\\
shafaei@cs.ubc.ca, little@cs.ubc.ca}
}


%


\maketitle

\begin{abstract}
Commonly used human motion capture systems require intrusive attachment of markers that are visually tracked with multiple cameras.
In this work we present an efficient and inexpensive solution to markerless motion capture using only a few Kinect sensors.
Unlike the previous work on 3d pose estimation using a single depth camera, we relax constraints on the camera location and do not assume a co-operative user.
We apply recent image segmentation techniques to depth images and use curriculum learning to train our system on purely synthetic data.
Our method accurately localizes body parts without requiring an explicit shape model.
The body joint locations are then recovered by combining evidence from multiple views in real-time.
We also introduce a dataset of \texttildelow{}6 million synthetic depth frames for pose estimation from multiple cameras and exceed state-of-the-art results on the Berkeley MHAD dataset.

\end{abstract}

\begin{IEEEkeywords}
depth sensors; human motion capture;

\end{IEEEkeywords}

%
\IEEEpeerreviewmaketitle

\section{Introduction}
\begin{figure}[!t]
\begin{center}
  \includegraphics[width=0.9\linewidth]{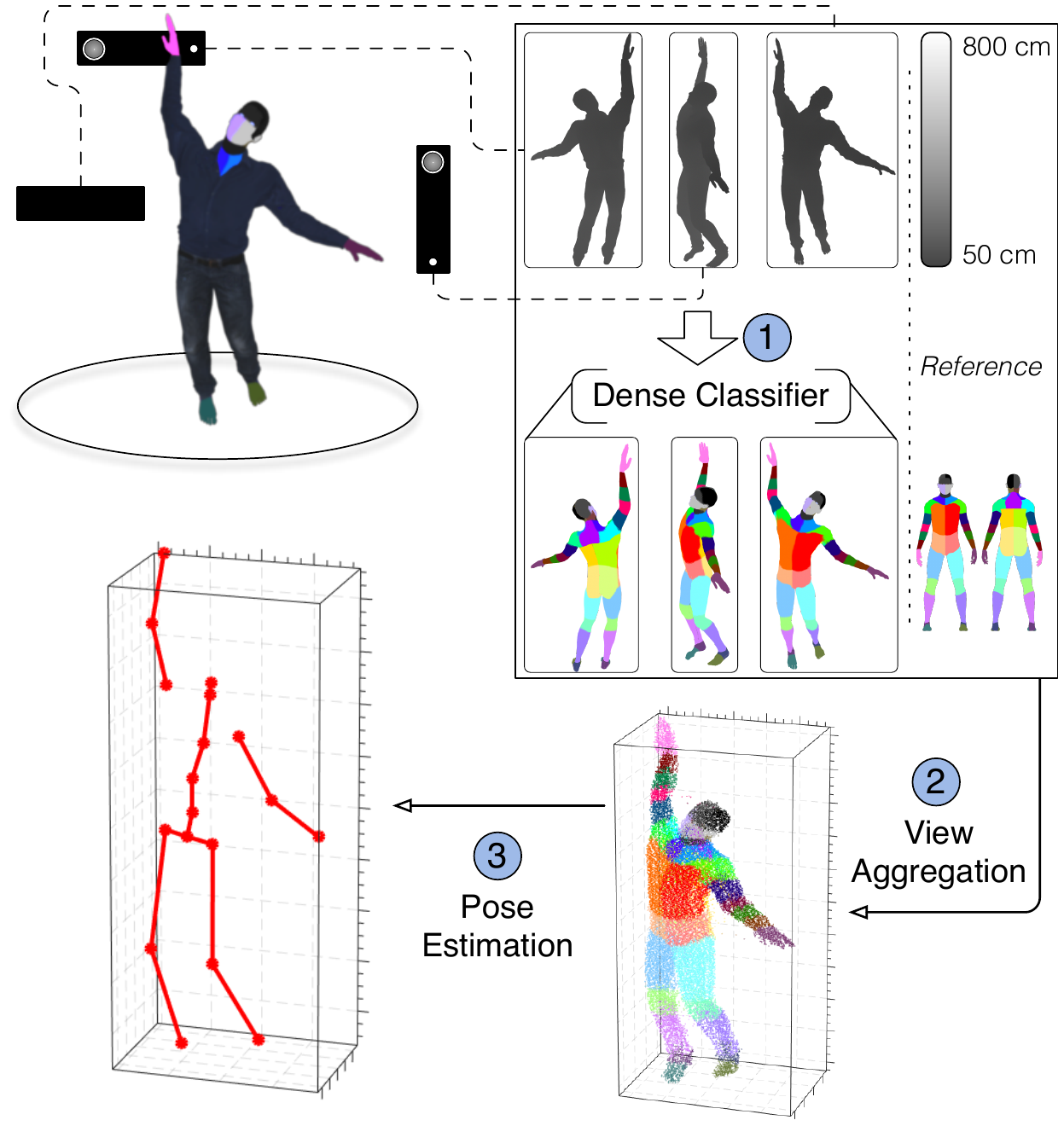}
\end{center}
  \caption{An overview of the problem and our pipeline. The output depth of the
  Kinect sensors is first passed through a dense classifier to identify the target
  body parts (1). We then merge all the cameras into a single 3d point cloud (2) and estimate the posture (3).}
  \label{fig:main_fig}
\end{figure}



Human motion capture is a process to localize and track the 3d location of body joints.
It is used to acquire a precise description of human motion which could be used for a variety of tasks such as
character animation, sports analysis, smart homes, human computer interaction, and health care.


In this work we are interested in extraction of 3d joint locations from multiple depth cameras.
Unlike previous work we do not impose contextual assumptions for limited scenarios such as home entertainment.
Instead, we focus on the problem of pose estimation to get comparable results to commercial motion capture systems.
Our target is to have a real-time, inexpensive, and non-intrusive solution to the general human motion capture problem.

While single view pose estimation has been extensively studied~\cite{ganapathi12realtime,shotton2013real,ye2014real,yub2015random},
surprisingly, pose estimation with multiple depth sensors is relatively unexplored.
One possible explanation is the challenge of overcoming the interference of multiple structured light sources.
However, with the recent adoption of time-of-flight sensors in the Kinect devices the interference has become unnoticeable.


One of the main challenges in this field is the absence of datasets for training or even a standard benchmark for evaluation.
At the time of writing the only dataset that provides depth from more than one viewpoint is the Berkeley Multimodal Human Action Database (MHAD)~\cite{ofli2013berkeley} with only two synchronized Kinect readings.

Our first contribution is a markerless human motion capture system.
Our pipeline uses an intermediate representation inspired by the work of Shotton \etal~\cite{shotton2013real}.
We split the multiview pose estimation task into three subproblems of (i) dense classification, (ii) view aggregation, and (iii) pose estimation (See Fig. \ref{fig:main_fig}).
Our approach can be distinguished from the previous work in several aspects. The context
of our problem is \textit{non-intrusive} and \textit{real-time}.
We do not assume \textit{co-operation}, a \textit{shape model} or a specific application such as home entertainment.
As a result our method is applicable to a wider range of scenarios. We only assume availability of \textit{multiple} externally calibrated depth cameras.

Our second contribution is three datasets of synthetic depth data for training multiview or single-view pose estimation systems.
Our three datasets contain varying complexity of shape and pose with a total of 6 million frames.
These datasets are constructed in a way that enables an effective application of curriculum learning~\cite{bengio2009curriculum}.
We show that a system trained on our synthetic data alone is able to generalize to real-world depth images.
We also release an open-source system for synthetic data generation to encourage and simplify further applications of computer generated imagery for interested researchers\footnote{All the data and source codes are accessible at the first author's homepage at \url{shafaei.ca}.}.

\section{Related Work}
\label{sec:rel_work}
The previous work on depth-based pose estimation can be categorized into two classes of top-down generative and bottom-up discriminative methods.


At the core of the generative methods there is a parametric shape model of the human body.
The typical approach is then to find the set of parameters that best describes the evidence.
Approaches to find the best parameters vary, but the common theme is to frame it as an expensive optimization problem~\cite{ganapathi2010real,ganapathi12realtime,helten2013personalization,ye2014real,Zhang2014leverage}.
Since a successful application of generative methods require reasonably accurate shape estimates, it is common practice to estimate the shape parameters beforehand~\cite{helten2013personalization,ye2014real}.
This dependence on the shape estimates limits the applicability of these methods to controlled environments with co-operative users.

Bottom-up discriminative models directly focus on the current input, usually down to pixel level classification to identify individual body parts.
For instance, Shotton \etal~\cite{shotton2013real} present a fast pipeline for single view pose estimation within a home entertainment context.
Their method classifies each pixel of a depth image independently using randomized decision forests.
The joint information is inferred by running a mean-shift algorithm on the classification output.
Girshick \etal~\cite{girshick2011efficient} improve the body estimates by learning a random forest to predict joint locations as a regression problem on the classification outputs directly.
It should be noted that neither the \textit{data} nor the \textit{implementation} of these methods are available for independent evaluation.
The popular Microsoft Kinect SDK only feeds data from the sensor directly.
More recently, Yub Jung \etal~\cite{yub2015random} presented a random-walk based pose estimation method that outperforms the previous state-of-the-art on a single view while performing at more than $1000\, \mathrm{fps}$.

In contrast, our solution is a middle ground between the discriminative and the generative approaches.
Although our approach does not require a parametric shape model, we integrate high-level spatial relationships of the human body during dense classification.
Specifically, by using convolutional networks one can even explicitly take advantage of CRF-like inference during classification to achieve smooth and consistent output~\cite{crfasrnn_iccv2015}.



\textbf{Multiview Depth}. Many of the existing methods for single view pose estimation, especially the top-down methods, can be naturally extended for the multiview case.
Michel \etal~\cite{michel2013tracking} show that through calibration of the depth cameras one can reconstruct a detailed 3d point cloud of  target subjects to do offline pose estimation by applying particle swarm optimization methods over the body parameters of a puppet.
The shape parameter of each subject is manually tuned before the experiments. Phan and Ferrie~\cite{phan2015towards} use optical flow in the RGB domain together with depth information to perform multiview pose estimation within a human-robot collaboration context at the rate of $8\,\mathrm{fps}$.
They report a median joint prediction error of approximately $15\,\mathrm{cm}$ on a T-pose sequence.
Zhang \etal~\cite{Zhang2014leverage} combine the depth data with wearable pressure sensors to estimate shape and track human subjects at $6\,\mathrm{fps}$. In this work we demonstrate how the traditional building blocks can be effectively merged to achieve the state-of-the-art performance in multiview depth-based pose estimation.

\textbf{Dense Image Classification}. Most current competitive approaches make use of deep convolutional networks~\cite{chen2014semantic,hariharan2015hypercolumn,long2015fully,Papandreou2015,crfasrnn_iccv2015}.
Long \etal~\cite{long2015fully} introduced the notion of fully convolutional networks.
Their method applies deconvolution layers fused with the output of the lower layers in the network to generate fine densely classified output.
Zheng \etal~\cite{crfasrnn_iccv2015} show that it is also possible to integrate the mean-field approximation on CRFs with Gaussian pairwise potentials as part of a deep convolutional architecture.
This approach enables end-to-end training of the entire system to get state-of-the-art accuracy in image segmentation.
The takeaway message is that under certain assumptions and modelling constraints the machinery of convolutional networks can be interpreted as reasoning within the spatial domain on dense classification tasks.
This proves useful when there are long distance dependencies in the spatial domain, such as the case when it is necessary to infer the side of the body to produce correct class labels, something that can not be decided locally.

\textbf{Synthetic Data}.
Shotton \etal~\cite{shotton2013real} augment the training data with synthetic depth images for training and evaluation of their method.
They also note that synthetic data in this case can be even more challenging than real data~\cite{shotton2013real}.
Park and Ramanan~\cite{park2015articulated} synthesize image frames of video to improve hand pose estimation.
Rogez \etal~\cite{rogez2015first} use depth synthesis to estimate hands' pose in an egocenteric camera setting.
Gupta \etal~\cite{gupta20143dpose, Guptaetal14} use synthetically generated trajectory features to improve cross-view action recognition by feature augmentation.
Since we take a fully supervised approach, dense labels are required to train the part classifier.
The use of synthetic data saved us a huge data collection and labelling effort.

\textbf{Curriculum Learning}.
Bengio \etal~\cite{bengio2009curriculum} describe curriculum learning as a possible approach to training models that involve non-convex optimization.
The idea is to rank the training instances by their difficulty.
This ranking is then used for training the system by starting with the simple instances and then gradually increasing the complexity of the instances during the training procedure.
This strategy is hypothesized to improve the convergence speed and the quality of the final local minima~\cite{bengio2009curriculum}.
Our experiments suggest that a controlled approach to training deep convolutional networks can be \textit{crucial} for training a better model, providing an example of curriculum learning in practice.

\section{Synthetic Data Generation}
\label{sec:syn_data}
We build a pipeline to generate samples of synthetic depth images of human body shapes using a variety of poses and viewpoints.
Our sampling process is described in Alg.~\ref{alg:synth}.

\begin{algorithm}
\caption{Sample data}\label{alg:synth}
$\mathcal{C}$: Pool of characters.\\
$\mathcal{L}$: Range of camera locations.\\
$\mathcal{P}$: Pool of postures.\\
$n$: Number of cameras.
\begin{algorithmic}[1]
\Procedure{Sample}{$\mathcal{C}, \mathcal{L}, \mathcal{P}, n$}
\State $c$ $\sim \mathrm{Unif}(\mathcal{C})$ \Comment{select a random character}
\State $l_{1:n}$ $\sim \mathrm{Unif}(\mathcal{L})$ \Comment{select $n$ random locations}
\State $p$ $\sim \mathrm{Unif}(\mathcal{P})$ \Comment{select a posture}
\State $S \gets$ Render depth image and groundtruth
\State \textbf{return} $S$ \Comment{$S=\{(D_i, G_i)\}_{i=1}^n$}
\EndProcedure
\end{algorithmic}
\end{algorithm}

\begin{figure}
\centering
\includegraphics[width=0.8\linewidth]{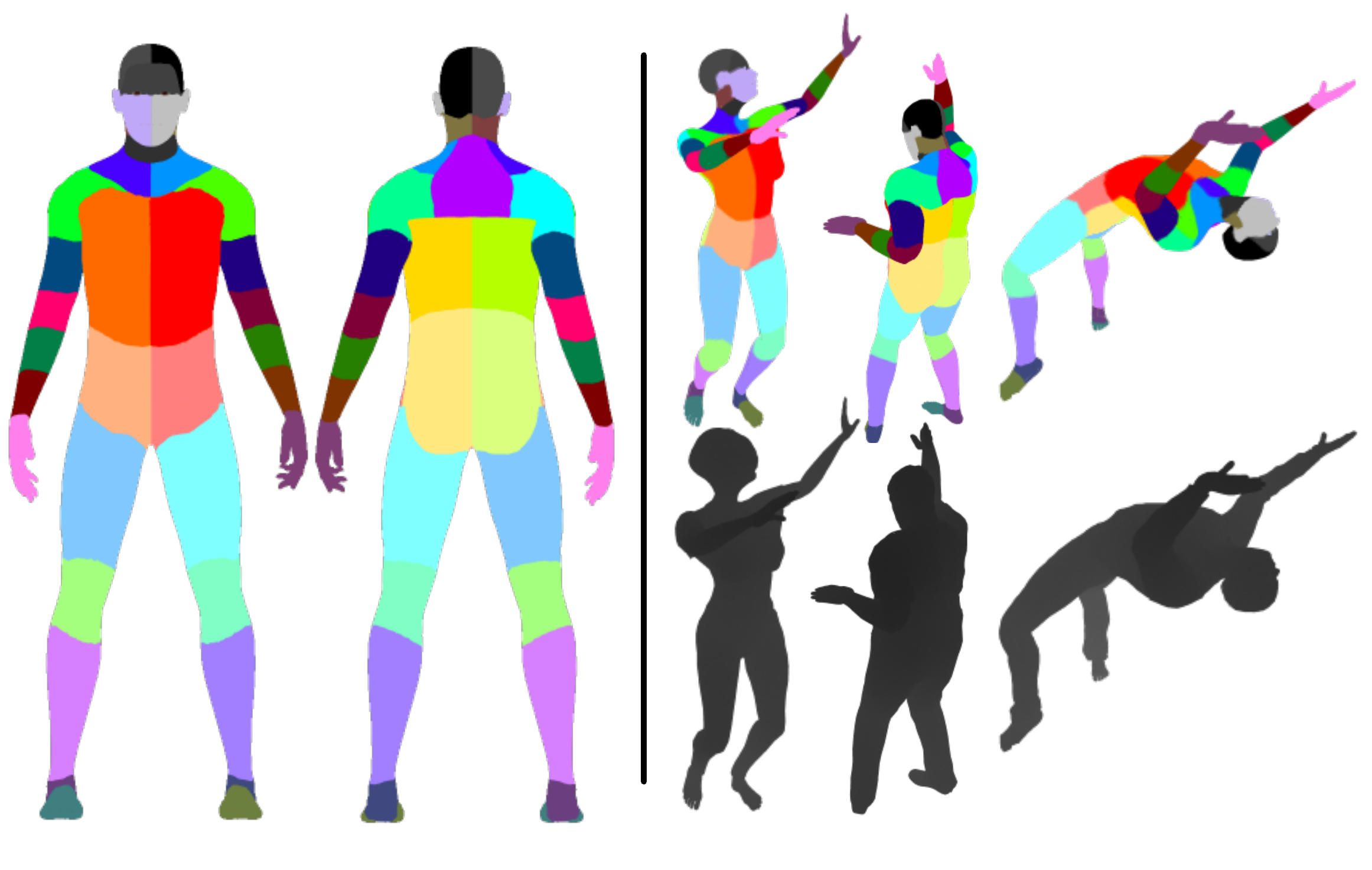}
\caption{Color-coded body texture to identify regions of interest (left), and two random samples of depth and groundtruth from our data (right).
        We define $43$ regions and distinguish left and right side of the body.}
\label{fig:body_regions}
\end{figure}

The character set $\mathcal{C}$ is the set of 3d models with variations in shape, age, clothing, and gender, generated using the free open-source project Make Human\footnote{\url{http://www.makehuman.org/}}.
We create a special skin texture for the characters to color-code each region of interest (see Fig. \ref{fig:body_regions}).
We found the $43$ body regions shown in Fig. \ref{fig:body_regions} to give sufficiently good results.

The camera location $\mathcal{L}$ is an interval description for camera placement.
The azimuth spans the $[-\pi, +\pi)$ range. The camera distance from the character is within $[1.5, 4]\, \mathrm{m}$.
Note that this distance constraint is only on the data generation process and not on the pose estimation pipeline.
We also need real human postures $\mathcal{P}$ to pose our characters.
We prepare this set by clustering all the human postures of the CMU motion capture\footnote{\url{http://mocap.cs.cmu.edu/}} dataset and picking the $\num{100000}$ most dissimilar poses.

The final parameter $n$ is the number of cameras for each sample.
The function \texttt{sample} returns a set of $n$ pairs $(D_i, G_i)$, where $D_i$ is the quantized depth image and $G_i$ is the groundtruth image as seen from the $i$-th camera.
While the depth image \textit{includes} the clothing items, in the groundtruth image we only render the textured skin.
Rendering is done with Autodesk Maya\footnote{\url{http://www.autodesk.com/products/maya/}}.
We set the camera parameters such that the generated data resembles the Kinect~2 output.


We define three datasets with varying complexity as our curriculum for training, as described in Table~\ref{tab:datasets}.
The easiest dataset has only one character with the subset of postures that are labeled with `walk' or `run' (\texttt{Easy-Pose}).
We then increase the difficulty by varying the set of possible postures (\texttt{Inter-Pose}), requiring the classifier to handle larger pose variations.
The final dataset includes the set of all characters with different physical shapes as well as all the postures from our dataset (\texttt{Hard-Pose}).
A transition from \texttt{Inter-Pose} to \texttt{Hard-Pose} would require learning shape variations.
The entire data is generated with $n=3$ cameras and each dataset has \texttt{train}, \texttt{validation}, and \texttt{test} sets with mutually disjoint postures. The datasets and the Python script to render customized data are made publicly accessible\footnote{Available at \url{shafaei.ca}.}.

\begin{table}[]
\centering
\caption{We generate three datasets as our curriculum. There are a total of $\num{100000}$ different postures.
The simple set is the subset of postures that have the label `walk' or `run'. Going from the first dataset
to the second would require pose adaptation, while going from the second to the third dataset requires
shape adaptation.}
\renewcommand{\arraystretch}{1.1}
\begin{tabular}{@{}llll@{}}
    \toprule
 Dataset                 & Postures & Characters & Samples \\
    \midrule
 Easy-Pose      & simple (\texttildelow{}{}$10\,\textrm{K}$)    & 1 & $1\,\textrm{M}$ \\
 Inter-Pose     & all ($100\,\textrm{K}$)                & 1 & $1.3\,\textrm{M}$ \\
 Hard-Pose      & all ($100\,\textrm{K}$)                & 16 & $300\,\textrm{K}$\\
    \bottomrule
\end{tabular}

\label{tab:datasets}
\end{table}

\section{Multiview Pose Estimation}
\label{sec:mul_pose}
Our framework consists of three stages (see Fig.~\ref{fig:main_fig}).
In the first stage we process the input of each sensor independently to generate an intermediate representation.
Each depth image is passed through a Convolutional Neural Network (CNN) to generate densely classified depth.
Having an intermediate representation gives us the flexibility to \emph{add} and \emph{remove} sensors from the environment with minimal reconfiguration.
To merge all the data we use the extrinsic camera parameters and reconstruct a 3d point cloud in the reference space.
This 3d point cloud is then labeled with the classification probabilities and used for pose estimation.
In the following sections we look at each step in detail.

\subsection{Dense Depth Classification}
\label{sec:dense_depth}
We use CNNs to generate densely classified depth.
Our architecture is motivated by the work of Long \etal~\cite{long2015fully}.
More specifically we use deconvolution outputs fused with the information from the lower layers to generate fine densely classified depth.
With this approach we are taking advantage of the information in the neighboring pixels to generate densely classified depth.
This is in contrast to random forest based methods such as \cite{shotton2013real} where each pixel is evaluated independently. 

\begin{figure*}
\begin{center}
  \includegraphics[width=0.85\linewidth]{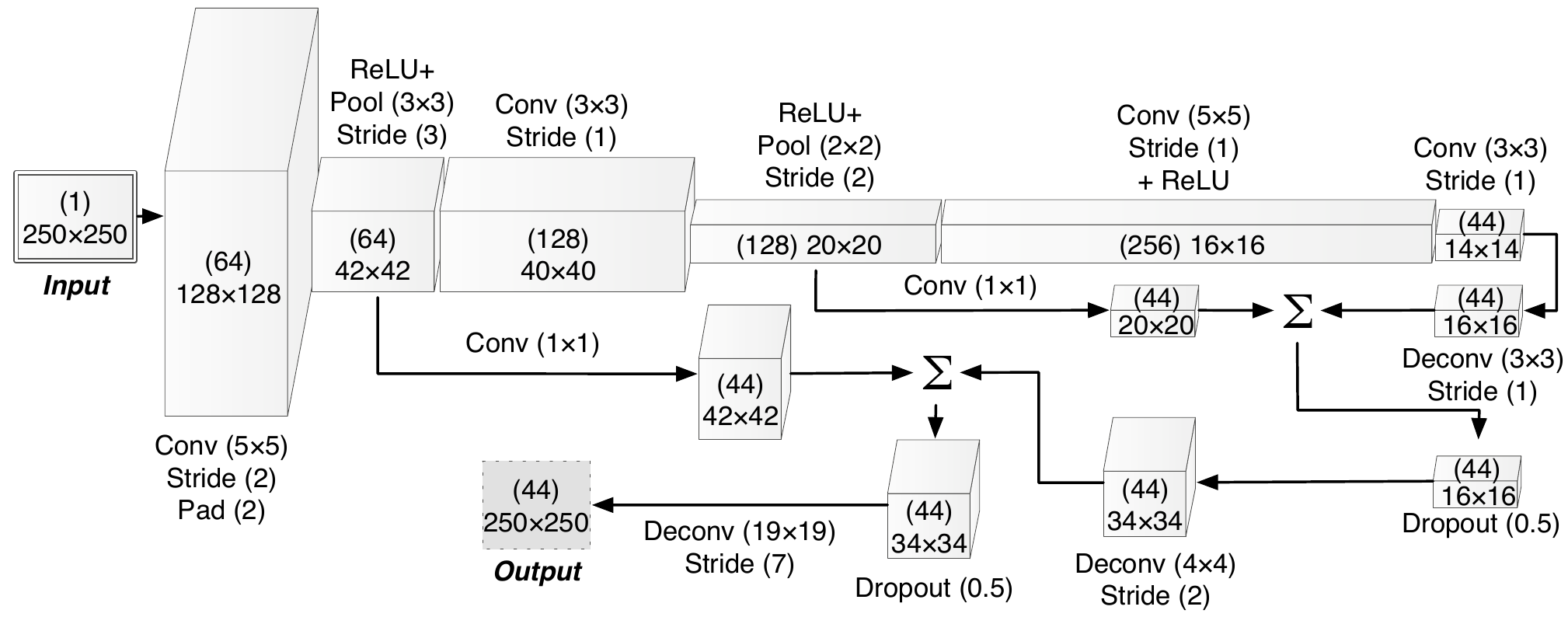}
\end{center}
  \caption{Our CNN dense classifier. The input is a $250\times{}250$ normalized depth image.
  The first part of the network generates a $44\times{}14\times{}14$ coarse densely classified depth
  with a high stride. Then it learns deconvolution kernels fused with the information from lower layers
  to generate fine densely classified depth. Like \cite{long2015fully} we use summation and crop alignment to fuse information.
  The input and the output blocks are not drawn to preserve the scale of the image.
  The number in the parenthesis within each block denotes the number of the corresponding channels.}
\label{fig:cnn}
\end{figure*}

\textbf{Preprocessing}. The input to our network is a $250\times{}250$~pixel depth
image with $30$ pixels of margin. The depth values are quantized to $255$ levels spanning the $[50, 800]\,\mathrm{cm}$ range.
We translate the average depth of each person to approximately $160\,\mathrm{cm}$ and spatially rescale to fit inside our window while preserving the aspect ratio.

\textbf{Classification}. The preprocessed image is then fed to the CNN of
Fig. \ref{fig:cnn}. The output of the network is $250\times{}250\times{}44$, representing
a $44$ dimensional vector of probabilities on each pixel for all the 43 classes and the background.
The deconvolution kernel size of $19\times{}19$ at the final stage enforces the spatial dependency between the adjacent pixels.

\textbf{Training}.
We initially attempted to train our network directly on \texttt{Hard-Pose}, however, in all of the trials with different optimization settings the accuracy of the network did not go above $50\%$ in average per-class classification.
Resorting to the curriculum learning idea of Bengio \etal~\cite{bengio2009curriculum} we simplified the task by defining easier datasets that we call \texttt{Easy-Pose} and \texttt{Inter-Pose} (see Table \ref{tab:datasets}).

We start training the network with the \verb'Easy-Pose' dataset.
Similar to Long \etal~\cite{long2015fully} we learned during our experiments that the entire network could be trained densely \textit{end-to-end} from \textit{scratch} without any class balancing schemes.
Each iteration processes eight depth images and we stop the initial phase of training at $250\,\mathrm{K}$ iterations reaching dense classification accuracy of $87.8\%$ on \texttt{Easy-Pose}.
We then fine-tune the resulting network on \texttt{Inter-Pose}, initially starting at an accuracy of $78\%$ and terminating after $150\,\mathrm{K}$ iterations with an accuracy of $82\%$.
Interestingly the performance on the \texttt{Easy-Pose} is preserved throughout this fine-tuning.
Finally we start fine-tuning on the \texttt{Hard-Pose} dataset and stop after $88\,\mathrm{K}$ iterations.
Initially this network evaluates to $73\%$ and by the termination point we have an accuracy of $81\%$.
In all of the above steps we test on the \texttt{validation} sets of each dataset as the \texttt{test} sets are reserved for the final pose estimation task.
The evolution of our three networks is shown in Table \ref{tab:net_prog}.
\begin{table}[]
\centering
\caption{The dense classification accuracy of the trained networks on the \texttt{validation} sets of the corresponding datasets. \texttt{Net~2} and \texttt{Net~3} are initialized
with the learned parameters of \texttt{Net~1} and \texttt{Net~2} respectively.}
\renewcommand{\arraystretch}{1.1}
\begin{tabular}{@{}lcccccc@{}}
    \toprule
    Dataset & \multicolumn{2}{c}{Easy-Pose} & \multicolumn{2}{c}{Inter-Pose} & \multicolumn{2}{c}{Hard-Pose}\\
    \cmidrule(lr){2-3}\cmidrule(lr){4-5}\cmidrule(lr){6-7}
    & Start & End & Start & End & Start & End \\
    \midrule
    Net 1 & $0\%$   & $87\%$   & --     & --        & --    & -- \\
    Net 2 & $87\%$  & $87\%$   & $78\%$ & $82\%$    & --    & -- \\
    Net 3 & $87\%$  & $85\%$   & $82\%$ & $79\%$    & $73\%$& $81\%$ \\
    \bottomrule
\end{tabular}
\label{tab:net_prog}
\end{table}
Notice how the final accuracy improved from $50\%$ to $81\%$ by applying curriculum learning.
The transition of training from \texttt{Net~1} to \texttt{Net~2} demands generalization of posture, while the transition from \texttt{Net~2} to \texttt{Net~3} requires shape invariance.
Our experiments demonstrate a real application of curriculum learning in practice.

\subsection{View Aggregation}
At this step we have collected $n$ densely classified outputs from our cameras.
We wish to generate a single labeled 3d point cloud.
Using the intrinsic parameters of each Kinect we can reconstruct a local point cloud.
We can merge all the point clouds using the extrinsic camera parameters to transform each point cloud to a reference space.

We then collect a set of statistics, namely:
(i) the median of each dimension, 
(ii) the covariance matrix,
(iii) the eigenvalues of the covariance matrix, 
(iv) the standard deviation of each dimension, 
and (v) the minimum and the maximum of each dimension 
for each class and use it as features for the later stage.
Since we are interested in real-time performance we limit ourselves to features that can be extracted quickly.
The redundancy of information in the feature space, such as including the eigenvalues of the covariance matrix as well as the covariance itself, is to ensure linear models can take advantage of all the information that is available.
We can devise filtering approaches for the merge process or perform temporal smoothing when the data is a sequence.
However, we have found that our computationally inexpensive approach is sufficient to achieve state-of-the-art results.
Our simple approach gives a feature vector $f \in \mathds{R}^{1032}$, that is, concatenation of all 24 features per class.

\subsection{Pose Estimation}
We treat the problem of pose estimation as regression.
For each joint $j$ in our target skeleton we would like to learn a function
$F_j(\cdot)$ that predicts the location of joint $j$ based on the feature vector $f$.
After examining a few real-time performing design choices such as \emph{linear regression} and \emph{neural networks} we learned that simple linear regression gives the best trade-off between complexity and performance.
We experimented with various neural network architectures, but the simple linear regression always performed better \emph{on average}.
Our linear regression is a least squares formulation with an $\ell_2$ regularizer which is also known as Ridge Regression and has a closed-form solution.
We also experimented with the LASSO counterpart to obtain sparser solutions but the improvements were negligible while
the optimization took substantially more time.
If the input data is over a sequence, we further temporally smooth the predictions by calculating a weighted average over the previous estimates, \ie, $\tilde{Y_i} = \sum_{j=0}^{K} \lambda_j Y_{i-j} \; \textrm{s.t.} \; \sum_{j=0}^{K} \lambda_j = 1$.
The regularizer hyper-parameters and the optimal smoothing weights are chosen automatically by cross-validation over the training data.




\section{Evaluation and Discussion}
\label{sec:eval}
In this section we present our evaluation results on three datasets. Since each dataset has a specific definition of joint locations we only need to train the regression part of our pipeline on each dataset.
The CNN depth classifier of Sec. \ref{sec:dense_depth} is trained \textit{once} and \textit{only} on the synthetic data.
Our experiments are run with Caffe~\cite{jia2014caffe} on a Tesla K40 GPU.
Each forward pass of the CNN takes $6.9\,\mathrm{ms}$ on the GPU or $40.5\,\mathrm{ms}$ on the CPU and requires $25\,\mathrm{MB}$ of memory per frame.
The entire pipeline can operate at $30\,\mathrm{fps}$ on a single machine while communicating with up to four Kinects.

\textbf{Evaluation Metrics}. There are two evaluation metrics that are commonly used for pose estimation: mean joint prediction error, and mean average precision at threshold.
Mean joint prediction error is the measure of average error incurred at prediction of each joint location.
Mean average precision at threshold is the fraction of predictions that are within the threshold distance of the groundtruth.
Because of the errors in groundtruth annotations it is also common to just report the mean average precision at $10\;\mathrm{cm}$ threshold~\cite{ye2014real,ganapathi12realtime,yub2015random}.

\textbf{Datasets}. We provide quantitative evaluations on the Berkeley MHAD~\cite{ofli2013berkeley} and our synthetic dataset.
Although for \textit{single view} depth images there are a few datasets such as EVAL~\cite{ganapathi12realtime}, and PDT~\cite{helten2013personalization},
 the only publicly available dataset for multiview depth at the moment of writing is the Berkeley MHAD~\cite{ofli2013berkeley}.

\subsection{Evaluation on UBC3V Synthetic}
For evaluation we use the \texttt{Test} set of the \texttt{Hard-Pose}.
This dataset consists of $\num{19000}$ postures with $16$ characters from three cameras placed at random locations.
These $\num{19000}$ postures are not present in the training set of our dataset and have not been observed before.
The groundtruth and the extrinsic camera parameters come from the synthetic data directly and there
are no errors associated with them.
Having groundtruth for body part regions and the posture helps us separate the evaluation of the dense classifier and the pose estimation technique.
That is, we can evaluate the pose estimation technique assuming perfect dense classification is available separately from the case where classification comes from our CNN.
This separation gives us insight on how improvement on dense classification is likely to affect pose estimation, and whether one should spend time on improving the classifier or the pose estimator.

For training we have the multi-step fine-tuning procedure as described in Sec. \ref{sec:dense_depth}.
We refer to the final fine-tuned network as \texttt{Net 3} throughout our experiments.

\begin{figure}[t]
\begin{center}
  \includegraphics[width=0.8\linewidth]{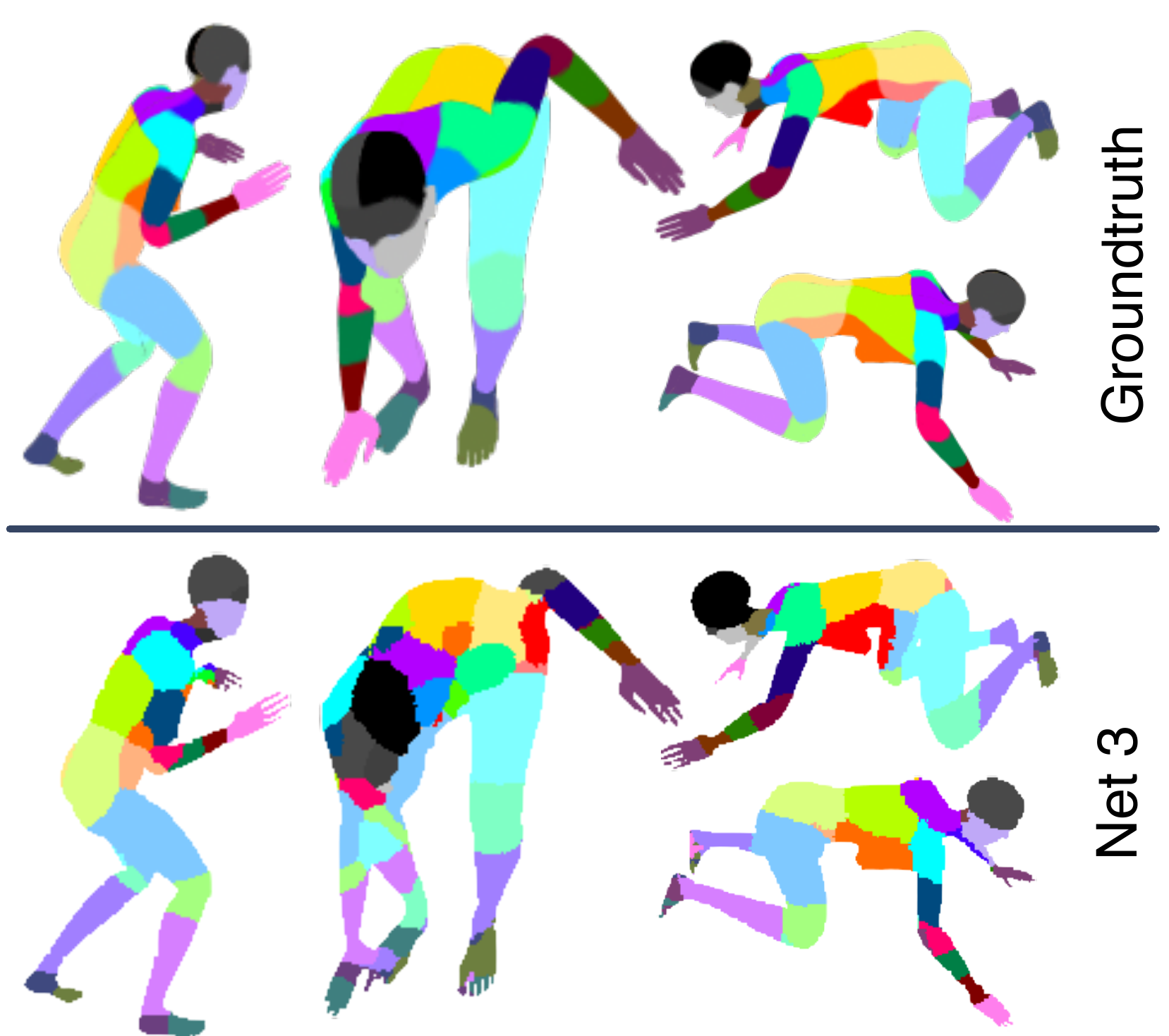}
\end{center}
  \caption{The groundtruth body part regions versus the output of \texttt{Net 3} classifier on the \texttt{Test} set of the \texttt{Hard-Pose}. Each pixel of the \texttt{Net 3} output is the color of the most likely class.}
\label{fig:dens_gt}
\end{figure}

\textbf{Dense Classification}. The \texttt{Test} set of the \texttt{Hard-Pose} includes $\num{57057}$ depth frames with dense class annotations that are synthetically generated.
Figure \ref{fig:dens_gt} demonstrates a few random classification samples and the corresponding groundtruth.
The CNN \textit{correctly} identifies the direction of the body and generates true classes for the \textit{left} and the \textit{right sides}, but seems to be ignoring sudden discontinuities in the depth.
For instance in the middle column of Fig.~\ref{fig:dens_gt} parts of the right shoulder are mixed with the head classes.
The overall accuracy of \texttt{Net 3} on the \texttt{Test} set is $80.6\%$, similar to the reported accuracy on the \texttt{Validation} set in Table~\ref{tab:net_prog}.

\begin{figure}[t]
\begin{center}
\includegraphics[width=0.8\linewidth, trim=3.7cm 8.9cm 3.9cm 9.3cm, clip=true]{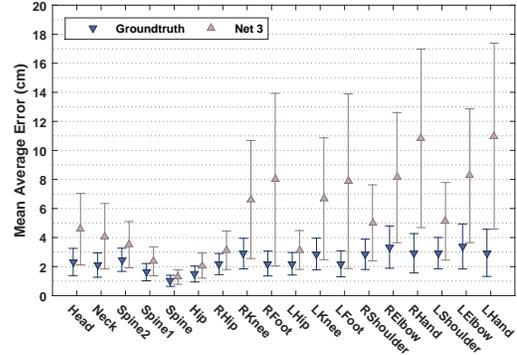}
\end{center}
  \caption{Mean average joint prediction error on the groundtruth and the \texttt{Net 3} classification output.
            The error bar is one standard deviation.
            The average error on the groundtruth is $2.44\,\mathrm{cm}$, and on the \texttt{Net~3} is $5.64\,\mathrm{cm}$.}
\label{fig:gt_mea}
\label{fig:net3_mea}
\end{figure}

\begin{figure}[t]
\begin{center}
\includegraphics[width=0.8\linewidth, trim=4.7cm 9cm 4.9cm 9.5cm, clip=true]{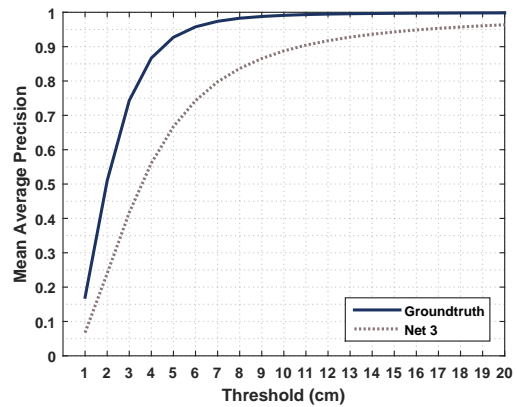}
\end{center}
  \caption{Mean average precision of the groundtruth dense labels and the \texttt{Net 3} dense classification output with accuracy at threshold $10\,\textrm{cm}$ of $99.1\%$ and $88.7\%$ respectively.}
\label{fig:gt_map}
\label{fig:net1_map}
\end{figure}

\textbf{Pose Estimation}. We evaluate our linear regression on the groundtruth class and the classification output of our CNN.
The estimate derived from the groundtruth serves as a lower bound on the error for the pose estimations algorithm.
The mean average joint prediction error is shown in Fig. \ref{fig:gt_mea}.
Our system achieves an average pose estimation error of $2.44\,\mathrm{cm}$ on groundtruth, and  $5.64\,\mathrm{cm}$ on the \texttt{Net~3}.
The gap between the two results is due to dense classification errors.
This difference is smaller on easy to recognize body parts and gets larger on the hard to recognize classes such as hands or feet.
It is possible to reduce this gap by using more sophisticated pose estimation methods at the cost of more computation.
In Fig.~\ref{fig:net1_map} we compare the precision at threshold.
The accuracy at $10\,\mathrm{cm}$ for the groundtruth and the \texttt{Net~3} is $99.1\%$ and $88.7\%$ respectively.


\subsection{Evaluation on Berkeley MHAD}
\begin{figure*}
\begin{center}
      \begin{subfigure}{0.74\linewidth}
          \includegraphics[width=\linewidth, trim=3.5cm 9.7cm 3.2cm 9.6cm, clip=true]{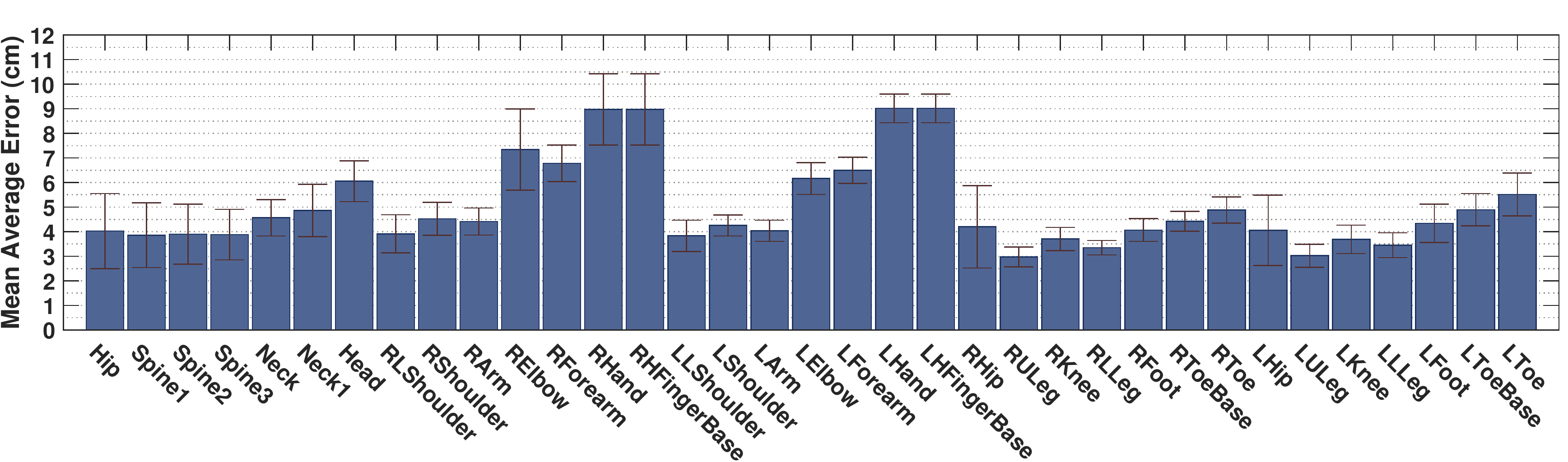}
          \caption{}
          \label{fig:mhad_joint}
      \end{subfigure}
      \hfill
      \begin{subfigure}{0.25\linewidth}
          \includegraphics[width=\linewidth, trim=4.4cm 8.90cm 4.9cm 9.5cm, clip=true]{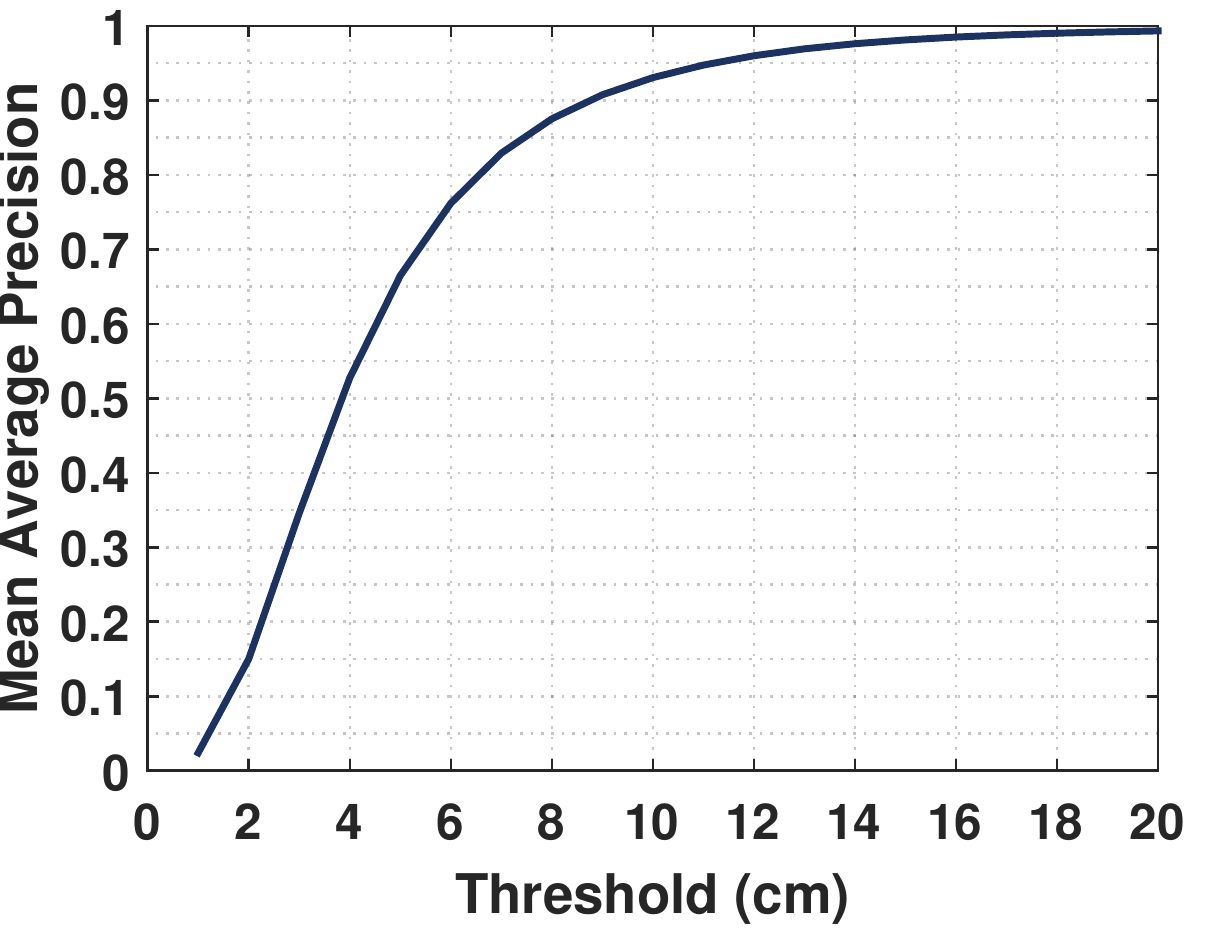}
          \caption{}
          \label{fig:mhad_acc}
      \end{subfigure}
\end{center}
  \caption{(a) Joint prediction mean average error on the Berkeley MHAD~\cite{ofli2013berkeley} dataset. The error bar is one standard deviation. The total mean average error per joint is $5.01\,\mathrm{cm}$.
            (b) Mean average precision at threshold for the entire skeleton on the Berkeley MHAD dataset.}
\end{figure*}

This dataset includes 12 subjects performing 11 actions while being recorded by 12 cameras, two Kinect ones, an Impulse motion capture system, four microphones, and six accelerometers.
The motion capture sequence with 35 joints is the groundtruth for pose estimation on this dataset (for a list of joints see Fig. \ref{fig:mhad_joint}).
We only use the depth information of the two opposing Kinect ones for pose estimation.

At the moment of writing there is no protocol for evaluation of pose estimation techniques on this dataset.
The leave-one-out approach is a common practice for single view pose estimation.
However, each action has five repetitions and we can argue in general that it may not be a fair indicator of the performance because the method can adapt to the shape of the subject of the test on other sequences to get a better result.
Furthermore, we are no longer restricted to only a few sequences of data as in previous datasets.

To evaluate the performance on this dataset we take the harder leave-one-subject-out approach, that is, for evaluation on each subject we train our system on all the other subjects.
This protocol ensures that no extra physical information is leaked during the training and can provide a measure of robustness to shape variation.

\begin{figure}[t]
\begin{center}
\includegraphics[width=\linewidth]{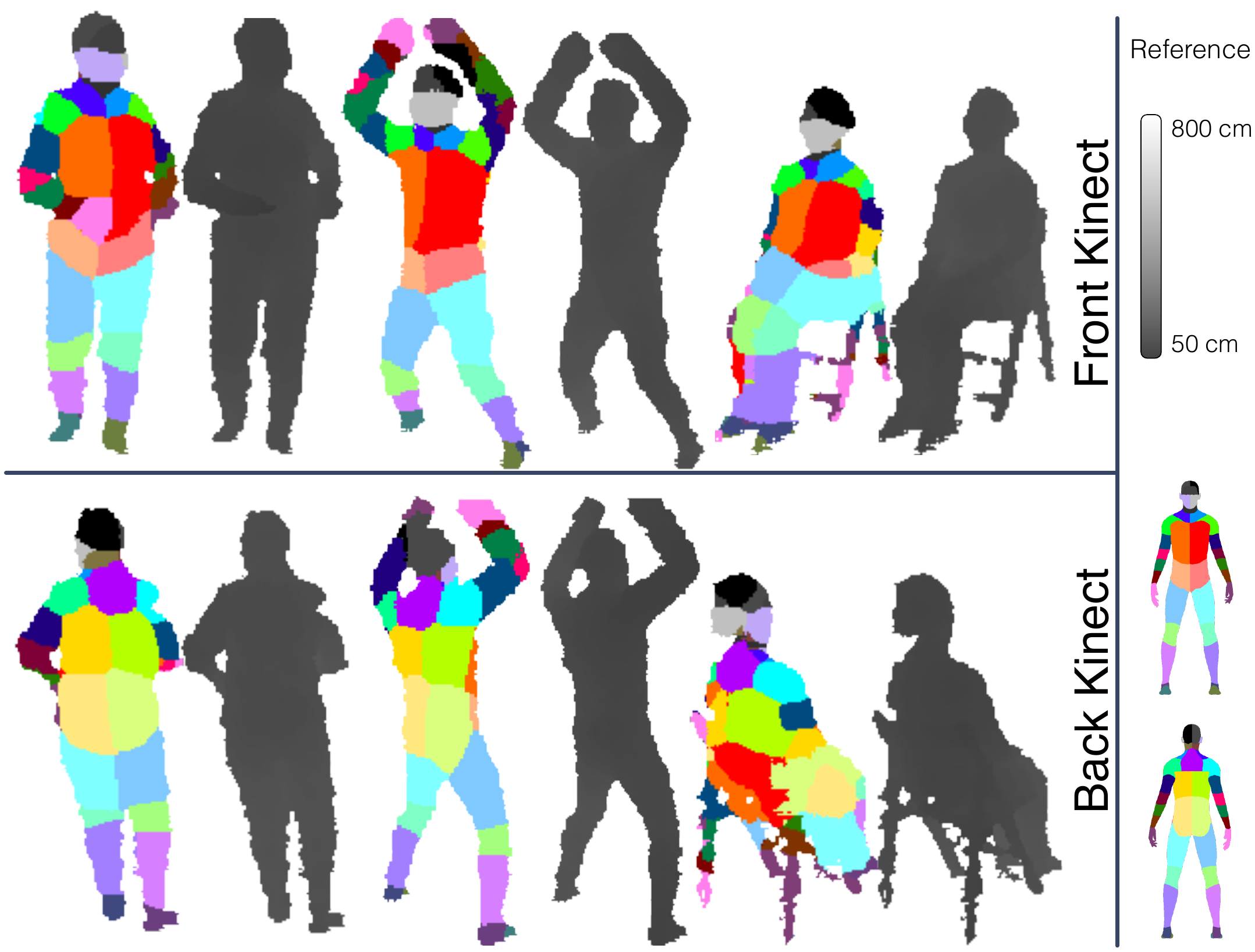}
\end{center}
  \caption{The dense classification result of \texttt{Net~3} together with the original depth image on the Berkeley MHAD~\cite{ofli2013berkeley} dataset.
            \texttt{Net~3} has been trained only on synthetic data. Each pixel is colored according to the most likely class.}
\label{fig:mhad_samples}
\end{figure}

\begin{figure}[t]
\begin{center}
\includegraphics[width=\linewidth]{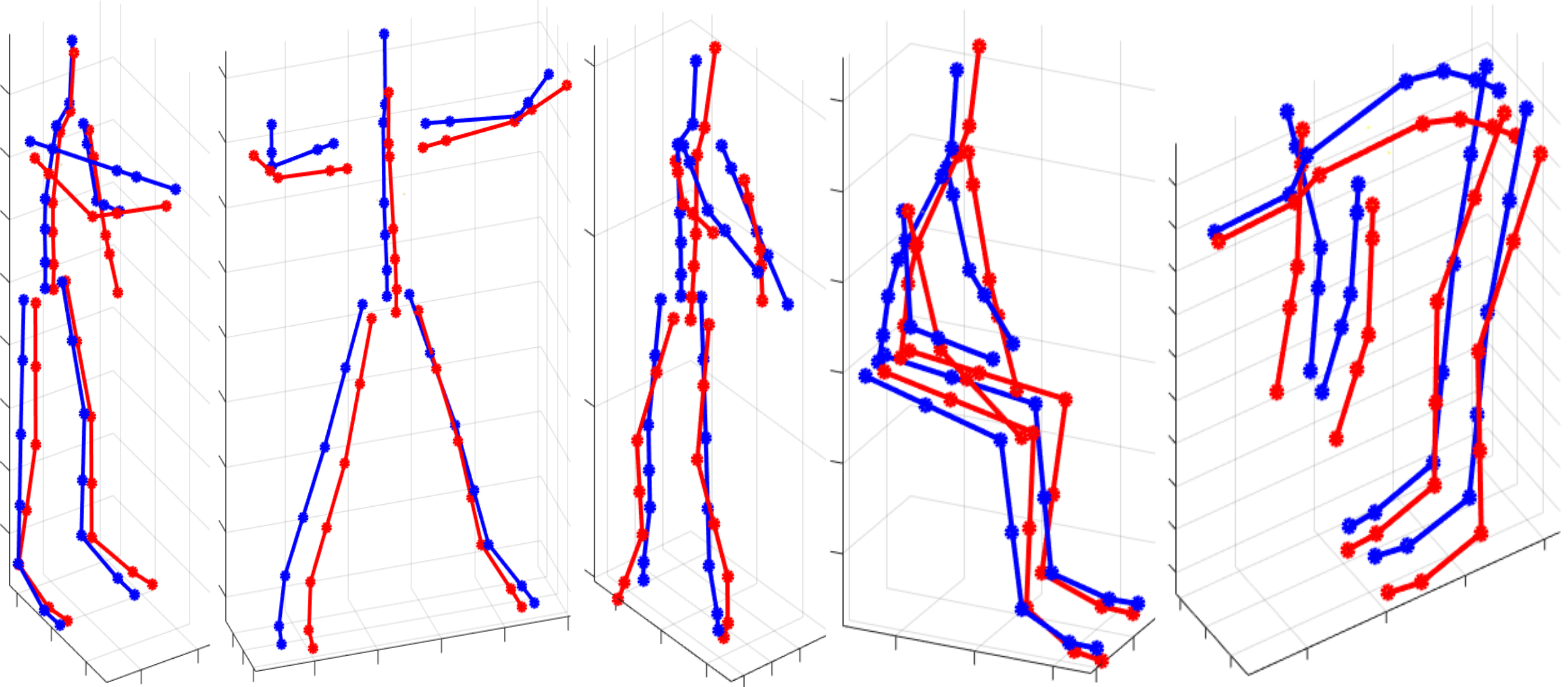}
\end{center}
  \caption{The blue color is the groundtruth of Berkeley MHAD~\cite{ofli2013berkeley} and the red color is our pose estimate.}
\label{fig:mhad_sample_pose}
\end{figure}

\textbf{Dense Classification}. To use the CNN that we have trained on the synthetic data we need to rescale the depth images of this dataset to match the output scale of Kinect~2 sensors.
After this step we simply feed the depth image to the CNN to get dense classification results. 
Figure \ref{fig:mhad_samples} shows the output of our dense classifier from the two Kinects on a few random frames.
Even though the network has been only trained on synthetic data, it is \textit{generalizing} well on the real test data.
As demonstrated in Fig. \ref{fig:mhad_samples}, the network has also successfully captured the long distance spatial relationships to correctly classify pixels based on the orientation of the body.
The right column of Fig. \ref{fig:mhad_samples} shows an instance of high partial classification error due to occlusion.
On the back image, the network mistakenly believes that the chair legs are the subject's hands.
However, once the back data is merged with the front data we get a reasonable estimate (see Fig. \ref{fig:mhad_sample_pose}).

\textbf{Pose Estimation}.
We use the groundtruth motion capture joint locations to train our system.
For each test subject we train our system on the other subjects' sequences.
The final result is an average over all the test subjects.

Figure~\ref{fig:mhad_joint} shows the mean average joint prediction error.
The total average joint prediction error is $5.01\,\textrm{cm}$.
The torso joints are easier for our system to localize than hands' joints, a similar behavior to the synthetic data results.
However, it must be noted that even the groundtruth motion capture on smaller body parts such as hands or feet is biased with a high variance.
During visual inspection of Berkely MHAD we noticed that on some frames, especially when the subject bends over, the location of the hands is outside of the body point cloud or even outside the frame, and clearly erroneous.
The overall average precision at $10\,\mathrm{cm}$ is $93\%$. 

An interesting observation is the similarity of performance on Berkeley MHAD data and the synthetic data in Fig. \ref{fig:net3_mea}.
This suggests that the synthetic data is a reasonable proxy for evaluating performance, as has been suggested by Shotton \etal~\cite{shotton2013real}.
Figure \ref{fig:mhad_acc} shows the accuracy at threshold for joint location predictions.

We also compare our performance with Michel \etal~\cite{michel2013tracking} in Table \ref{fig:comp_mich}.
Since they are using an alternative definition of skeleton that is derived by their shape model,
we only evaluate over a subset of the joints that are closest with the locations presented in Michel \etal~\cite{michel2013tracking}.
Note that the method of \cite{michel2013tracking} uses predefined shape parameters that are optimized for each subject \textit{a priori} and does not operate in real-time.
In contrast, our method does not depend on shape attributes and operates in real-time.
Following the procedure of \cite{michel2013tracking} we evaluate by testing the subjects and testing the actions.
Our method improves the previous mean joint prediction error from $3.93$ to $3.39$ ($13\%$) when tested on subjects
and $4.18$ to $2.78$ ($33\%)$ when tested on actions.


\begin{table}[]
\centering
\setlength{\tabcolsep}{0.15em}
\caption{Mean and standard deviation of the prediction error by testing on subjects and actions with the joint definitions of Michel \etal~\cite{michel2013tracking}.
        We also report and compare the accuracy at $10\,\mathrm{cm}$ threshold.}
\renewcommand{\arraystretch}{1.1}
\begin{tabular}{@{}lcccccc@{}} \toprule
    & \multicolumn{3}{c}{Subjects} & \multicolumn{3}{c}{Actions}\\
    \cmidrule(lr){2-4}\cmidrule(lr){5-7}
    & Mean & Std & Acc (\%) & Mean & Std & Acc (\%) \\
    \midrule
    OpenNI~\cite{michel2013tracking}          & $5.45$     & $4.62$ & $86.3$      & $5.29$ & $4.95$ & $87.3$\\
    Michel \etal~\cite{michel2013tracking}    & $3.93$     & $2.73$ & $96.3$      & $4.18$ & $3.31$ & $94.4$  \\
    Ours                                  & $\textbf{3.39}$     & $\textbf{1.12}$ & $\textbf{96.8}$      & $\textbf{2.78}$ & $\textbf{1.5}$  & $\textbf{98.1}$  \\
    \bottomrule
\end{tabular}
\label{fig:comp_mich}
\end{table}

\section{Conclusion}
\label{sec:conc}
We presented an efficient and inexpensive markerless motion capture system that uses only a few Kinect sensors.
Our system only assumes availability of calibrated depth cameras and is capable of real-time performance without requiring an explicit shape model or cooperation.
We further presented a dataset of \texttildelow{}6 million synthetic depth frames for pose estimation from multiple cameras.
Our experiments demonstrated an application of curriculum learning in practice and our system exceeded state-of-the-art multiview pose estimation performance on the Berkeley MHAD dataset.


\section*{Acknowledgements}
We would like to thank Ankur Gupta for helpful comments and discussions. We gratefully acknowledge the support of NVIDIA Corporation with the donation of the GPUs used for this research. This work was supported in part by NSERC under Grant CRDPJ 434659-12 and the ICICS/TELUS People \& Planet Friendly Home Initiative at UBC.

{\small
\bibliographystyle{IEEEtran}
\bibliography{IEEEabrv,egbib}
}

\newpage
\section*{Supplementary Material}
\subsection{Evaluation on EVAL}
\begin{figure}[t]
\begin{center}
\includegraphics[width=\linewidth]{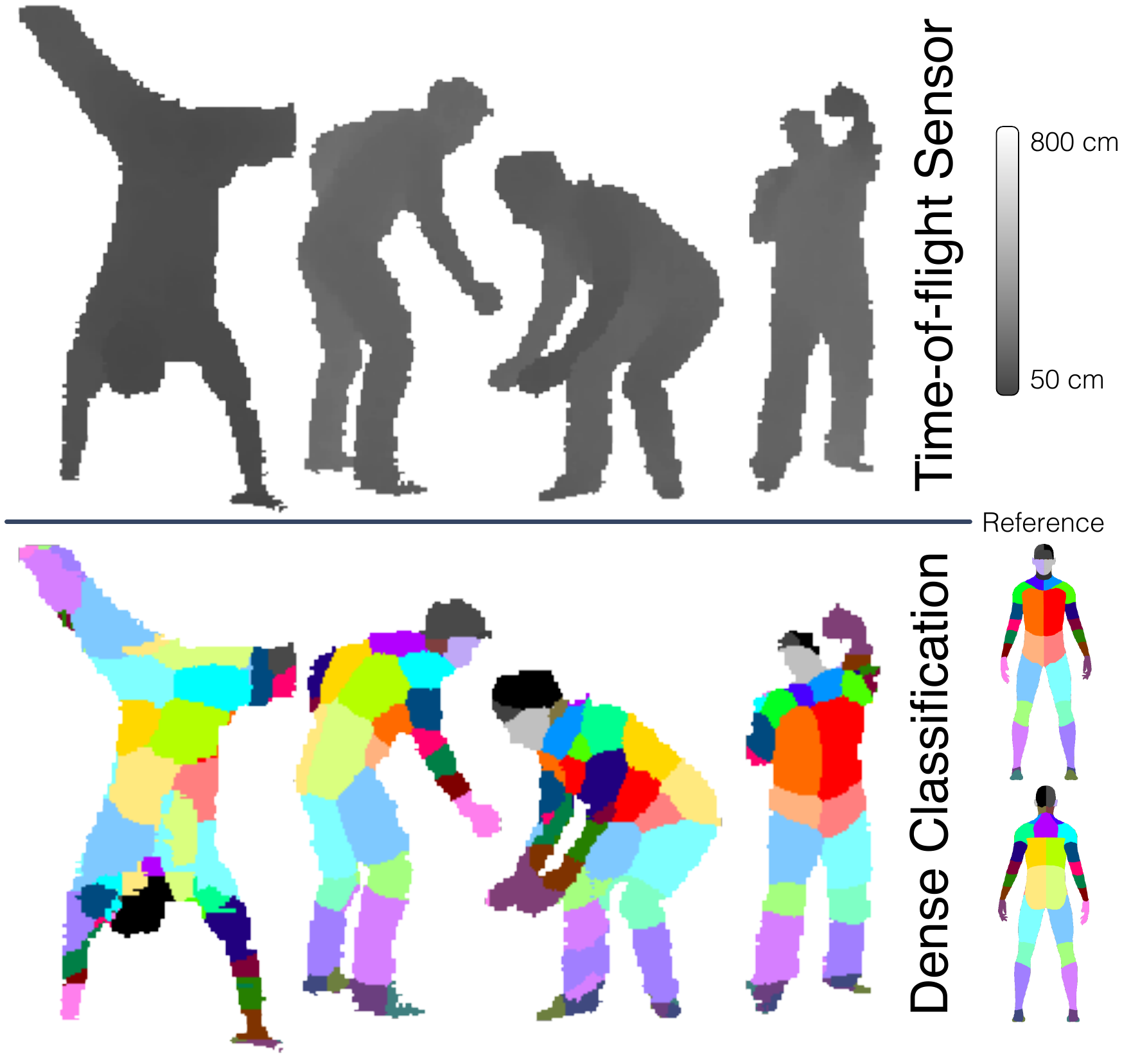}
\end{center}
  \caption{The classification result of \texttt{Net~3} and the original depth image on the EVAL~\cite{ganapathi12realtime} dataset.
            Each pixel is colored according to the most likely class.
            }
\label{fig:eval_samples}
\end{figure}

There are a total of 24 sequences of 3 subjects with 16 joints.
To generate a depth image from this dataset we must project the provided point cloud into the original camera surface and then rescale the image to resemble the Kinect two output.
Figure \ref{fig:eval_samples} shows four random dense classification outputs from this dataset.
The first column of Fig. \ref{fig:eval_samples} shows an instance of the network failing to confidently label the data with front or back classes, but the general location of torso, head, feet and hands is correctly determined.
The accuracy of our preliminary results suggests that single depth pose estimation techniques can benefit from using the output of our dense classifier.
Due to the invisibility of almost half of the body regions in the single view regime, a high percentage of our constructed features will not be available for evaluation.
While our system can make reasonable predictions with only a few missing features in multiview settings, it does not handle single-view pose estimation well.  

\end{document}